\documentclass[runningheads]{templates/eccv26/llncs}

\usepackage{templates/eccv26/eccv}

\usepackage{templates/eccv26/eccvabbrv}
\usepackage{graphicx}
\usepackage{booktabs}
\usepackage[accsupp]{axessibility}
\usepackage[pagebackref,breaklinks,colorlinks,citecolor=eccvblue]{hyperref}

\usepackage{orcidlink}

\newcommand{\papertitleshort}{LCV: 
Directorial Video Control using Spatial Blocking}

\newcommand{\papertitle}{\name: \\
Directorial Video Control using Spatial Blocking}

\usepackage{xspace}
\newcommand{\name}{{LooseControlVideo}\xspace}
\newcommand{\abbrname}{{LCV}\xspace}

\makeatletter
\DeclareRobustCommand\onedot{\futurelet\@let@token\@onedot}
\def\@onedot{\ifx\@let@token.\else.\null\fi\xspace}

\makeatother

\usepackage{xcolor}
\usepackage{wrapfig}
\definecolor{light-gray}{gray}{0.95}

\usepackage{graphicx}
\usepackage{booktabs}

\usepackage{booktabs}
\usepackage{multirow}
\usepackage{array}
\usepackage{makecell}
\usepackage{tabularx}
\usepackage{colortbl}
\usepackage{xcolor}
\usepackage{dsfont}

\newcommand{\tabsmall}{\fontsize{8}{9}\selectfont}

\newcommand{\up}{\,$\uparrow$}
\newcommand{\down}{\,$\downarrow$}

\newcommand{\best}[1]{\textbf{#1}}
\newcommand{\second}[1]{\underline{#1}}

\begin{document}

\title{\papertitle}
\titlerunning{\papertitleshort}

\author{Shariq Farooq Bhat
\and
Niloy J. Mitra \and
Kalyan Sunkavalli}

\authorrunning{S.~F.~Bhat et al.}

\institute{Adobe Research\\
\url{https://shariqfarooq123.github.io/LooseControlVideo/}
}

\maketitle

\begin{abstract}
  Precise 3D spatial orchestration in text-to-video generation remains a significant challenge, particularly for multi-object scenes where semantic layout and temporal dynamics are often entangled. While existing depth-conditioned models achieve good structural fidelity, they necessitate dense, frame-accurate guidance that is labor-intensive to author for dynamic events involving deformable objects. We present \name~(\abbrname), a framework that enables intuitive and expressive control by using sparse, oriented 3D boxes as a ``blocking'' proxy. This allows users to author high-level layout and trajectory while leveraging a video generative model to generate realistic occlusions, dynamics and interactions. We achieve this by fine-tuning a Wan 2.2 backbone on a video dataset annotated with DNOCS, a novel encoding for 3D size, orientation and depth-ordered occlusions.
  Furthermore, our method allows for localized refinement—such as adjusting a jump trajectory or adding an interaction—with minimal disruption to the global scene context. Extensive evaluations on the nuScenes, HO-3D, and BEHAVE benchmarks demonstrate that \abbrname significantly outperforms existing 2D-box and flow-based baselines. Our findings indicate a 1.2-3x improvement in Trajectory Error; 2x improvement in Rigid Motion Consistency; and  a 1.5-2x increase in Occlusion Accuracy over current state-of-the-art layout-conditioned models, demonstrating that oriented 3D primitives provide good geometric prior for complex, multi-agent video authoring.


\end{abstract}


\section{Introduction}
\label{sec:intro}

While modern video diffusion models~\cite{veo,wan2025wan21,sora} have achieved remarkable photorealism, choreographing compelling narratives that involve intricate spatio-temporal synchronization and complex multi-object interactions remains an elusive task using current controls. 
This is because these models often lack the \textit{directorial} controls required to orchestrate specific physical events, leaving a significant gap between generative capability and creative intent.

\begin{figure}[t!]
    \centering
    \includegraphics[width=\linewidth]{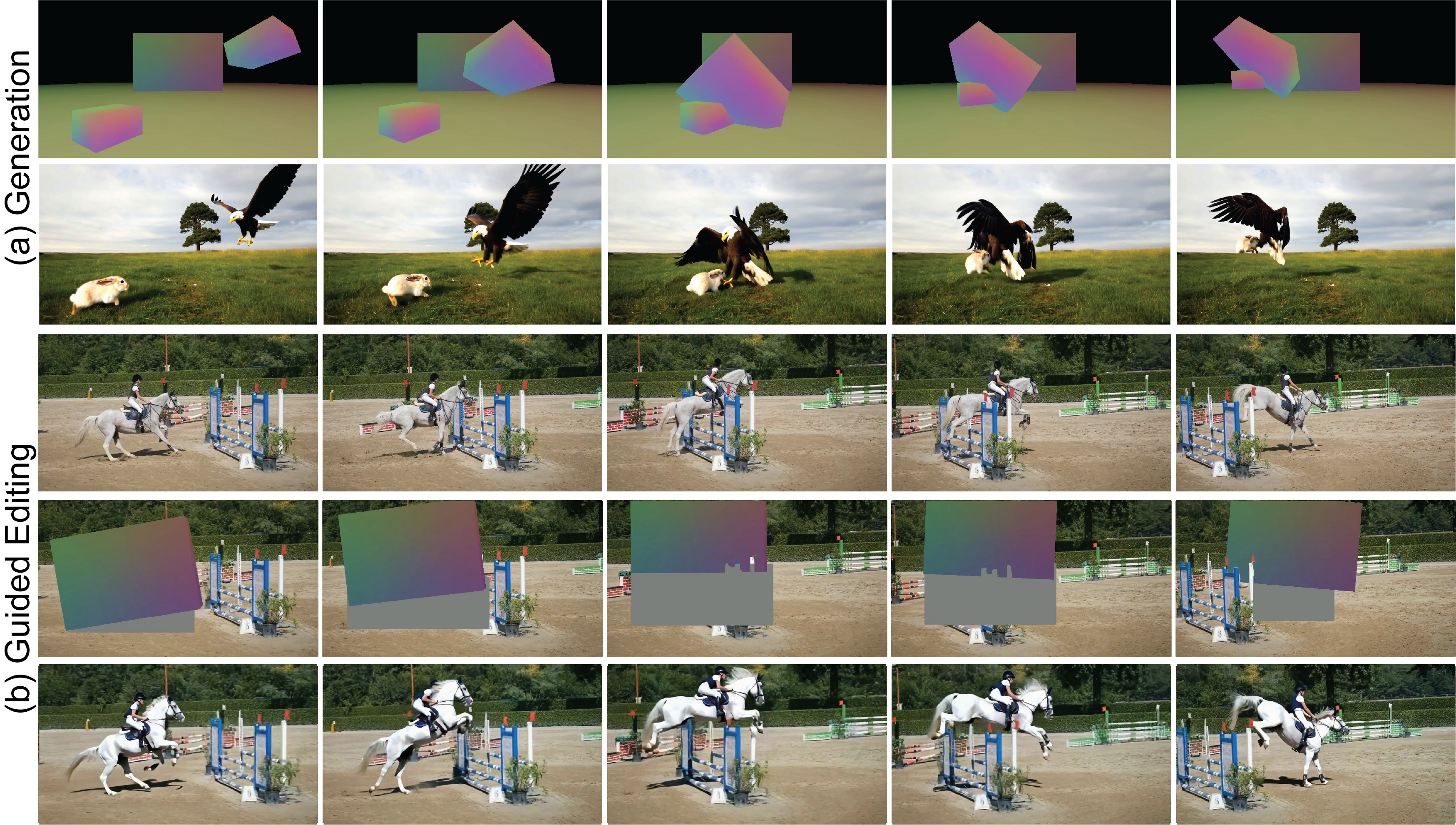}
    \caption{
    \textbf{Cinematic video authoring and editing via oriented 3D blocking.} Our method, \name, enables precise choreography of complex multi-object interactions using sparse, oriented 3D bounding boxes. \textit{Top:} From a coarse blocking of \textit{an eagle stooping to catch a rabbit}, our model infers realistic morphological deformations and temporal synchronization. \textit{Bottom:} \name facilitates intuitive motion editing; by lifting the jumping horse video (DAVIS dataset~\cite{davis-Perazzi2016}) into our oriented 3D proxy space, a user can easily modify the jump trajectory of the horse while preserving its visual identity and the scene's temporal consistency. See supplemental. 
    }
    \label{fig:teaser}
\end{figure}

Existing control modalities present a difficult trade-off: natural language is too imprecise to describe exact spatial trajectories, whereas dense video signals, such as per-frame depth or edge maps, are nearly impossible for a user to manually author for dynamic scenes. This difficulty stems from the fact that structure guidance (e.g., depth video) conflates two distinct control axes: (i)~the layout, motion and interactions of the camera and objects in the scene, and (ii)~the fine-grained object poses and deformations arising from those interactions. For instance, requiring a user to provide a frame-accurate depth sequence of an eagle stooping to catch a rabbit—capturing every wing-beat and skeletal contraction—is an impractical prerequisite.

Drawing inspiration from the ``blocking'' phase in professional cinematography, where directors orchestrate scene layouts using coarse spatial proxies to define movement, we introduce \name (LCV), a method that empowers users to author complex video events through the intuitive manipulation of oriented 3D primitives. Here, the user focuses solely on the high-level choreography (the layout and motion path and timing), while our generative model infers the secondary dynamics and realistic deformations (the execution). By allowing users to author these trajectories via simple 3D modeling tools or rigid-body physics simulations, \name effectively offloads the burden of geometric modeling as well as complex object deformations to the diffusion process, bridging the gap between casual authoring and cinematic precision.

We demonstrate that a simple-to-author sequence of \textit{oriented 3D boxes}, when rendered as a structured temporal signal, provides a sufficient geometric prior for fine-tuning a large-scale video generator to achieve precise motion control. We expect the users to only provide a (sparse) set of keyframes consisting of oriented 3D boxes rendered from the target camera’s perspective.
To resolve ambiguities in complex motion, we propose Depth-modulated Normalized Object Coordinate Space (DNOCS) representation, a novel coloring scheme to jointly encode local orientation as well as global depth.

To bridge the gap between coarse 3D proxies and high-fidelity synthesis, we render our 3D DNOCS representation into a 2D control signal (see \Cref{fig:teaser}). We demonstrate that this enriched representation provides a robust conditioning signal for the Wan 2.2 DiT architecture, allowing the model to learn the \textit{execution} of physical motion from sparse \textit{intent} signals. To facilitate training without manual 3D annotation, we propose an automated pipeline that extracts temporally-tracked 3D oriented boxes from large-scale in-the-wild video datasets. By utilizing this system, we create a training corpus of $\sim$10K aligned RGB-DNOCS video pairs, enabling the model to learn the relationship between 3D spatial occupancy and realistic object-level dynamics.

We demonstrate \name through quantitative and qualitative evaluations on datasets including nuScenes~\cite{nuscenes20} for navigation and HO-3D~\cite{hampali2020ho3d} and BEHAVE~\cite{bhatnagar22behave} for articulated interactions. We propose novel metrics for spatial blocking adherence, an area unaddressed in prior work. Our experiments show \name significantly outperforms state-of-the-art baselines; 
we achieve a 1.2-3$\times$ reduction in Trajectory Error and a 1.5-2$\times$ increase in Occlusion Accuracy while maintaining base DiT photorealism. Our contributions include:
(i)~a 3D-aware spatial blocking paradigm using oriented boxes to decouple choreography from local deformation;
(ii)~a specialized control mechanism and encoding for Diffusion Transformers enabling intuitive \textit{loose} control;
(iii)~an automated pipeline for lifting 2D video datasets into 3D-conditioned training sets; and
(iv)~metrics to evaluate video model adherence to spatial controls.

\section{Related Work}

\paragraph{Image edit metaphors and handles.} 
The rapid evolution of generative diffusion models has established a robust foundation for high-quality 2D image synthesis~\cite{ho2020denoising, rombach2022high, saharia2022photorealistic}, yet the usability of these systems hinges entirely on the specific \emph{modalities of control} exposed to the user. Early efforts in 2D spatial conditioning ranged from sparse inputs like 2D bounding boxes~\cite{li2023gligen} and sketches~\cite{voynov2023sketch} to dense signals such as segmentation maps~\cite{wang2022pretraining} and depth gradients~\cite{zhang2023adding, mou2023t2i}. ControlNet~\cite{zhang2023adding} and followups successfully unified these dense modalities (e.g., edge maps, human poses) for localized guidance; however, they often force users into \textit{indirect control}—where the complexity of authoring means that signals must be derived from existing imagery—making it difficult to create from scratch. 
Alternatives emerged by allowing \emph{text/instruction} based editors~\cite{InstructPix2Pix, LEDITSpp} that offer zero-shot convenience but lack precise compositional steering, while \emph{drag- and point-based} interfaces~\cite{DragDiffusion, DragAPart} enable local geometric manipulation but provide limited understanding of global articulation or temporal continuity. Coarse geometry-driven controls utilizing loose boxes~\cite{loosecontrol} reduced the need for exact shape guidance; they do not support temporal changes, especially arising from deformation and complex interactions. Our work addresses this critical gap; unlike 2D-centric methods~\cite{li2023gligen,loosecontrol} that offer loose control without 3D-awareness, we introduce oriented 3D primitives that combine the intuitive flexibility of dense/sparse keyframes with the loose 3D scene blocking, enabling the choreography of complex, deforming multi-object events.

\paragraph{Video generators and control paradigms} The transition from U-Net based inflated attention architectures~\cite{videodiffusionmodels, makeavideo} to  Diffusion Transformer (DiT)~\cite{peebles2023scalable} has catalyzed a surge in high-fidelity video generation. While proprietary models~\cite{sora,veo,gemini} have demonstrated unprecedented temporal consistency and scale, the community has seen a parallel rise in open-weight/source alternatives~\cite{zheng2024opensorademocratizingefficientvideo,yang2025cogvideox,wan2025wan21}. Despite these foundational advancements, steering these models remains primarily limited to high-level text instructions or global style descriptors. Efforts to port the spatial precision of ControlNet to the temporal domain, such ControlVideo~\cite{zhao2023controlvideoconditionalcontroloneshot} or Ctrl-V~\cite{luo2024ctrlvhigherfidelityvideo}, typically still rely on dense video signals (e.g., depth or Canny maps, or pixel-space boxes). However, as seen in recent work on cinematic steering~\cite{hu2024motionctrl, he2025cameractrl}, dense depth is not only difficult for users to author for complex dynamic scenes but also conflates camera ego-motion with local object deformation. While recent modules like CameraCtrl~\cite{he2025cameractrl} and Direct-a-Video~\cite{yang2024directavideo} introduce dedicated camera and trajectory parameters, they often struggle to maintain identity and structural integrity during intricate multi-object interactions. We leverage DiT architecture's capacity for long-context temporal reasoning, conditioning it on sparse oriented primitives that sidestep the need for authoring/sourcing dense, hard-to-obtain structural maps.

\paragraph{3D-Aware Video Synthesis/Editing.} Video control has increasingly moved toward 3D-aware scene composition, drawing inspiration from professional cinematography. Recent works have explored this through varied lenses: LLM-as-director frameworks~\cite{c3v, Lin2023VideoDirectorGPT,kizil2025lamplanguageassistedmotionplanning} translate high-level scripts into spatial coordinates for rigged shapes, while Gen3C~\cite{gen3c} and other world models focus on maintaining global 3D consistency. Parallel efforts like Diffusion-as-Shader~\cite{das} treat the diffusion process as a rendering pass over 3D tracked signals, often requiring a pipeline of rigging, animated meshes, and subsequent depth estimation to provide structure. In the realm of user-centric interaction, Boximator~\cite{wang2024boximator} introduced intuitive 2D box-guided control for object selection and motion, and recent point-track methods like Edit-by-Track~\cite{editbytrack} allow for precise motion editing via sparse point trajectories. However, these 2D-centric or track-based approaches face a dual challenge: (i)~authoring exhaustive 2D tracks/boxes for dynamic, deforming scenes remains difficult, and (ii)~they struggle to represent complex 3D behaviors such as axial spins, viewpoint-consistent deformations, and depth-ordered occlusions. By distinguishing between global rigid motion (encoded in our oriented 3D boxes) and local semantic deformation (inferred by the generator), we introduce a scalable oriented proxy that captures the full 6-DOF intent of an interaction without the overhead of full 3D rigging or dense tracking.

\section{Method}

\subsection{Overview}

We address the problem of controllable video generation and editing using light-weight 3D proxy signals. Our goal is to enable intuitive spatial control while preserving the generative flexibility of modern video diffusion models.

Let $y := \{y_t\}_{t=1}^{T}$ denote a target video of $T$ frames and let $p$ denote a text prompt describing the scene. We introduce control signal $b := \{b_t\}_{t=1}^{T}$ represented as a sequence of oriented 3D boxes. Each frame-level control map $b_t$ encodes the (time-varying) spatial layout of (multiple) objects in the scene. In addition, we allow the user to specify 3D cameras, static or dynamic, $c := \{c_t\}_{t=1}^{T}$ for the video to be generated. 

Our objective is to model the conditional distribution $p_\theta(y \mid p, b, c)$; this corresponds to video \textit{generation}, controlled by the input text $p$ and user-authored 3D bounding boxes $b$ as seen through the cameras $c$. In addition, we would also like to support video \textit{editing} scenarios where a (possibly spatially or temporally sparse) input source video, $v$, may be provided. In the most general setting, our video generator needs to model the distribution $p_\theta(y \mid p, b, c, v)$.

There are many design choices for how the bounding box and camera controls can be provided to a video generator.
A box at time $t$ can be parameterized by coordinate representation: center $\mathbf{o}_t \in \mathbb{R}^3$, scale $\mathbf{s}_t \in \mathbb{R}^3$, and rotation $\mathbf{R}_t \in SO(3)$.
A naive approach for conditioning a video diffusion model on 3D box and camera parameters is to encode these parameters directly using MLPs and append the resulting embeddings as special tokens into the attention layers of the base DiT. Variants of this idea can be constructed by modifying how the tokens are injected or fused with the backbone. However, such a design suffers from a fundamental flaw: \textit{video diffusion models operate primarily in two spatial dimensions and do not seem to possess an explicit strong understanding of 3D geometry~\cite{jeong2025track4genteachingvideodiffusion}}. As a result, a tokenized coordinate representation of the box and camera parameters provides only implicit geometric cues.

In particular, critical spatial cues, such as depth ordering, occlusion relationships, and perspective projection, are not directly observable from the raw coordinate representation of the boxes and camera parameters. 
This introduces a significant representation mismatch: the conditioning signal is expressed in abstract 3D parameters, while the base video model operates entirely in the 2D image domain. Bridging this gap requires the model to internally reason about complex 3D scene geometry, effectively reconstructing the rendering pipeline that maps 3D structure to 2D observations. This substantially increases the learning burden and makes the conditioning difficult to utilize. Indeed, in practice, we observe that models trained with such token-based conditioning frequently fail to converge reliably and often learn to ignore the control inputs altogether, producing generations that do not respect the intended 3D structure at all.

\begin{figure}[t!]
    \centering
    \includegraphics[width=\linewidth]{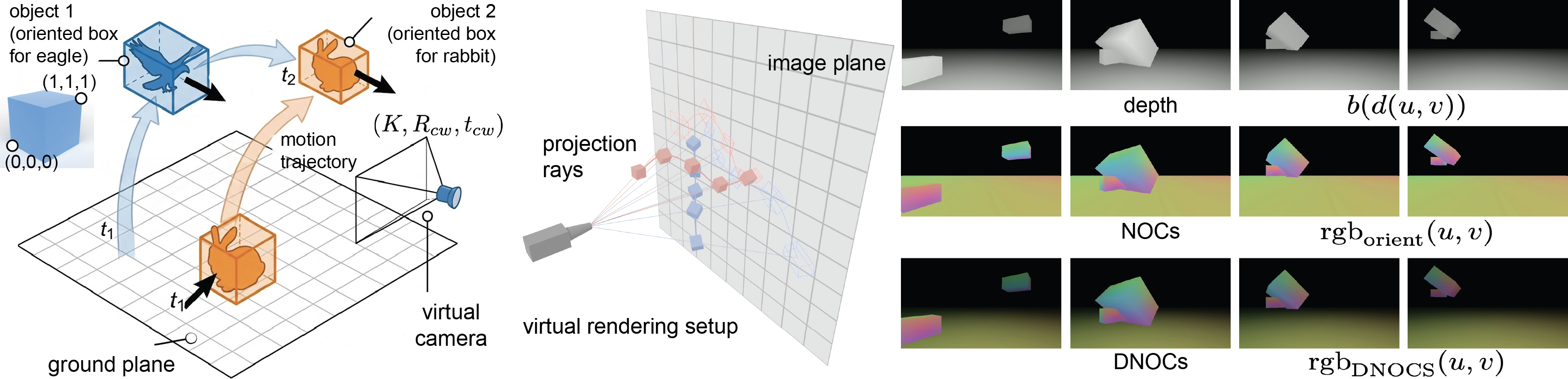}
    \caption{
   \textbf{Overview of the 3D control space and virtual rendering setup.} (Left) The control space abstractly defined as 3D oriented boxes and their motion trajectories over time ($[t_1,t_2]$) for distinct objects (e.g., eagle, blue and rabbit, orange) on an optional ground plane relative to a virtual camera. (Middle) The virtual rendering setup illustrates the projection of 3D object primitives onto a virtual image plane via projection rays originating from the camera. (Right) Illustrations of the dense maps rendered for a sequence of the moving objects: depth map, NOCs (Normalized Object Coordinates), and DNOCs (our depth+NOCs).
   }
    \label{fig:looseCntrl_illustration}
\end{figure}

The key insight of our work is to render the (time-varying) boxes ($b_t$) using the given camera ($c_t$) to a conditioning signal in image space \textit{before} feeding to the video model, aligning the control representation with the domain in which the base video model operates. By rendering beforehand, we effectively amortize the geometric projection step, exposing depth ordering, occlusion, and perspective cues explicitly rather than requiring the model to infer them from abstract parameters. This results in a control signal,
$v^{ctrl}$, that is expressive enough to enable complex spatio-temporal operations without requiring architectural modifications to pre-trained video models. Moreover, this results in a simple yet expressive control interface that can handle \textit{both} video generation and editing.



\subsection{DNOCS: Rendered Oriented 3D Box Representation}
Our goal is to render boxes such that the control signal can encode all possible object motion trajectories including 3D translations, rotations and scales, while respecting occlusions and relative relationships. We introduce Depth-modulated Normalized Object Coordinate Space (DNOCS), a hybrid color representation that jointly encodes local orientation as well as global depth. DNOCS makes key geometric cues such as object orientation, depth ordering, and occlusion relationships and object interactions directly observable in the conditioning frames. Inspired by NOCS~\cite{wang2019nocs} maps, we color each box using its local normalized coordinate frame, allowing the model to infer object orientation and relative pose. We render these colored boxes into the video frame using the specified camera, $c_t$. In addition, we  modulate the color intensity according to the object’s depth with respect to the camera, introducing a global distance cue that standard NOCS encodings lack. See \Cref{fig:looseCntrl_illustration} for an illustration. 

Given camera intrinsics $K$ and extrinsics $(R_{cw}, t_{cw})$ (world $\rightarrow$ camera), and a box defined by center $C_w$, rotation $R_{wb}$ (box $\rightarrow$ world), and half-extents $h$, we cast a ray for each pixel and compute its intersection with the box.
For pixel $(u,v)$ the camera-space ray direction is, 
\begin{equation}
d_c(u,v)=\Big(\tfrac{u-c_x}{f_x}, \tfrac{v-c_y}{f_y}, 1\Big).
\end{equation}
Transforming to world space using $R_{wc}=R_{cw}^\top$ gives, 
\begin{equation}
o_w=-R_{wc}t_{cw}, \qquad d_w(u,v)=R_{wc}d_c(u,v).
\end{equation}
We express the ray in the box-local frame using $R_{bw}=R_{wb}^\top$:
\begin{equation}
o_b=R_{bw}(o_w-C_w), \qquad d_b(u,v)=R_{bw}d_w(u,v).
\end{equation}
We then compute ray--box intersection using the standard slab test for ray--AABB intersection in the local frame $[-h, h]$, yielding a hit mask $m(u,v)$, intersection parameter $t(u,v)$, and local surface point, 
\begin{equation}
p_b(u,v)=o_b+t(u,v)d_b(u,v).
\end{equation}
Finally, we obtain the corresponding camera-space depth is as, 
\begin{equation}
p_w=o_w+t d_w, \qquad p_c=R_{cw}p_w+t_{cw}, \qquad
z(u,v)=(p_c)_z,
\end{equation}
with background pixels assigned $z=+\infty$.

Our DNOCS representation of control image combines a depth-independent orientation hue with a depth-dependent brightness, yielding a single 3-channel conditioning signal that carries both local orientation and global depth ordering.
We first convert the local intersection to a unit direction on the sphere,
\begin{equation}
n(u,v) = \frac{p_b(u,v)}{\lVert p_b(u,v)\rVert + \epsilon},
\end{equation}
and map directions to RGB with approximately constant luminance using a smooth spherical color wheel,
\begin{equation}
\mathrm{rgb}_{\text{orient}}(u,v)
= \mathrm{unit\_sphere\_color}\big(n(u,v); L, a\big),
\end{equation}
where $L$ is a fixed luminance (we use $L=0.55$) and $a$ controls chroma strength (we use $a=0.35$).
This yields easily distinguishable hues across orientations. Next, we compute a normalized inverse depth signal $d\in[0,1]$ (closer $\rightarrow 1$, farther $\rightarrow 0$) so nearer surfaces have larger values:
\begin{equation}
d(u,v) = 1 - \mathrm{clip}\!\left(\frac{z(u,v) - z_{\min}}{z_{\max}-z_{\min}},\,0,1\right),
\end{equation}
where $(z_{\min}, z_{\max})$ are $2^{nd}$ and $98^{th}$ percentile depths computed from $\{z(u,v)\mid m(u,v)=1\}$. Finally, we turn depth into a multiplicative brightness with an exponential falloff and a nonzero floor:
\begin{equation}
b\!\left(d\right)=\beta_{\min} + (1-\beta_{\min})\exp\!\big(-k(1-d)\big),
\end{equation}
with $\beta_{\min}=0.08$ and $k=2.0$ by default.
The final DNOCS rendering is
\begin{equation}
\mathrm{rgb}_{\text{DNOCS}}(u,v)=\mathrm{rgb}_{\text{orient}}(u,v)\odot b\big(d(u,v)\big),
\end{equation}
where $\odot$ denotes channel-wise multiplication (broadcasting $b$ over RGB channels).
Background pixels are set to black ($\mathrm{rgb}_{\text{hyb}}=0$) with $m=0$ and $z=+\infty$.




\paragraph{Representation of input videos for Editing.}
Our goal is to enable loose control for both video generation and editing. The latter requires specifying the boxes in the context of the input video. We achieve this by compositing the rendered DNOCS representation into the input video frames in an occlusion-aware manner to construct the complete control, $v^{ctrl}$, to the video generator. For compositing, we estimate depth of the input video using VideoDepthAnything\cite{video_depth_anything25}. We make the authoring of $v^{ctrl}$ as flexible as possible by allowing each frame of this context input to be any one of the following: (i)~DNOCS-only frames; (ii)~Input video segments to specify frames that must stay the same; (iii)~Completely black (empty) frames which video model must fill in; and (iv)~Spatially composited mixtures of input video frames and the DNOCS renderings. We also support gray masks over input video frames for regions that are to be removed. 

This unified formulation enables the same model to handle both generation and editing. Moreover, it is highly expressive and allows users to choose which form of control should be applied on a per-frame and per-object basis. \Cref{fig:teaser} shows control video examples for generation (top) and editing (bottom).

\subsection{Automatic Training Data Pipeline}

Training \name requires a dataset of videos annotated with boxes for a wide variety of objects in the scenes. We build these annotations for a curated subset of internal stock videos 
of approximately 10k real videos using an automated, robust pipeline. For each video, we obtain object tracks using GroundingDINO~\cite{liu2023grounding} and SAM-based segmentation~\cite{ravi2024sam} to produce per-frame object masks. We then estimate per-object point clouds using monocular depth prediction~\cite{video_depth_anything25} within the masked regions and fit oriented 3D bounding boxes per frame~\cite{li_globFit_sigg11}. 
We refine the raw box trajectories using 3D Kalman filtering procedure to produce temporally consistent box sequences.

\textit{Control Video Construction.}
During training, we randomly construct control signals by randomly choosing between box-only renders (70\%), and a mix of partial real video segments and the spatially composited mixtures (30\%). 

\subsection{Training}
We build on a ControlNet\cite{ControlNet}-style architecture, WAN-VACE~\cite{vace}, and keep the base video diffusion architecture unchanged. The control video, $v^{ctrl}$ is provided through the standard VACE conditioning pathway, which injects control residuals into the backbone DiT model. We LoRA fine-tune the VACE modules (rank 64) for 10k iterations on 4 H100 80GB GPUs while keeping the base model frozen. This design choice isolates the effect of the proposed control representation and demonstrates that strong controllability can be achieved without architectural modification. 

At inference time, the user can switch between generation and editing modes simply by changing how $v^{ctrl}$ is constructed: \textit{(i) Pure Generation.} When $v^{ctrl}$ consists of only DNOCS renderings, the model synthesizes a full video consistent with the specified camera and object layout and motion. 
\textit{(ii) Editing and Interpolation.} When $v^{ctrl}$ is a combination of the input video and the DNOCS renderings, the model preserves the visible regions while generating the remaining content consistent with the box controls and text prompt.

\begin{figure}[t!]
    \centering
    \includegraphics[width=\linewidth]{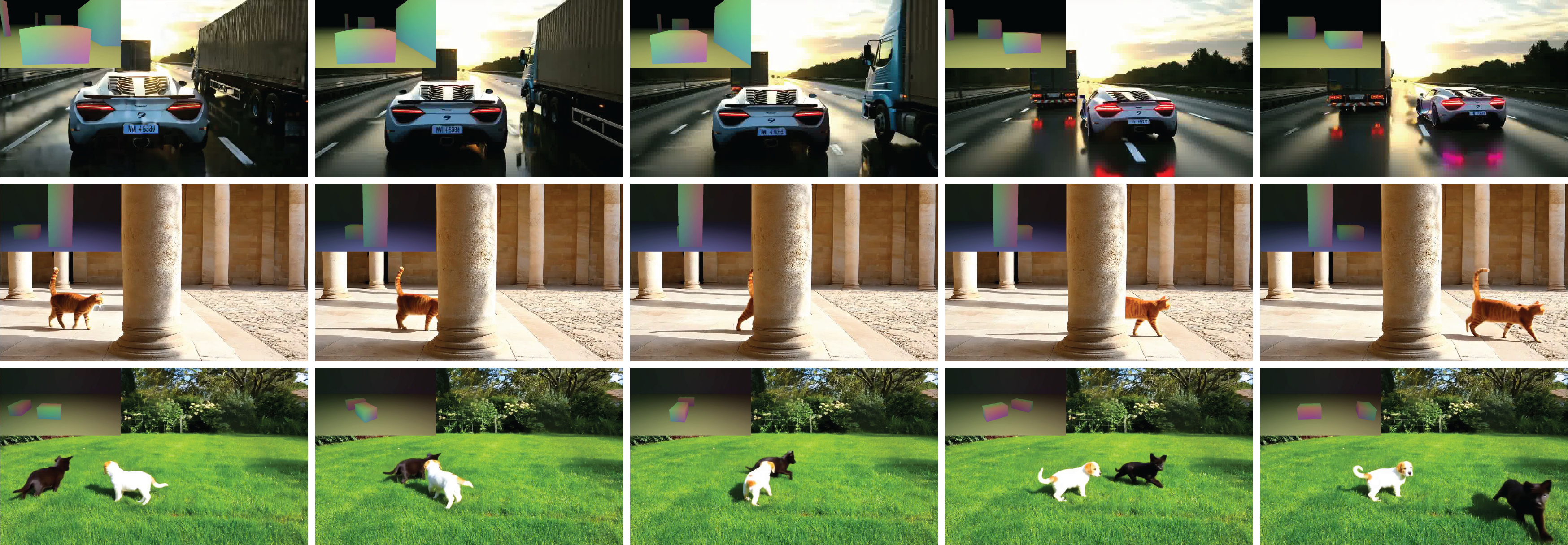} 
    \caption{
    \textbf{Video generation results.} Insets visualize the input 3D oriented box sequences used as conditioning proxies. 
    \textit{Top:} High-speed weaving and maneuvering. The oriented boxes guide a vehicle through narrow gaps between trucks, capturing subtle 6-DOF rotations and triggering responsive effects like brake light activation. 
    \textit{Middle:} Robust occlusion and shadow consistency. A cat maintains its identity and temporal coherence while walking behind a large pillar; note the preservation of intricate, viewpoint-consistent shadows and the smooth reappearance from total occlusion. 
    \textit{Bottom:} Articulated interaction and rhythmic motion. Two puppies interact with natural deformation and gait as one rotates around the other, demonstrating LCV's ability to infer complex morphological changes from sparse loose/rigid proxies. 
    In all cases, our easy-to-author oriented boxes anchor the structural orchestration required to drive the video generator. See supplemental webpage for videos and more examples.}
    \label{fig:results_gallery}
\end{figure}

\begin{figure}[t!]
    \centering
    \includegraphics[width=\linewidth]{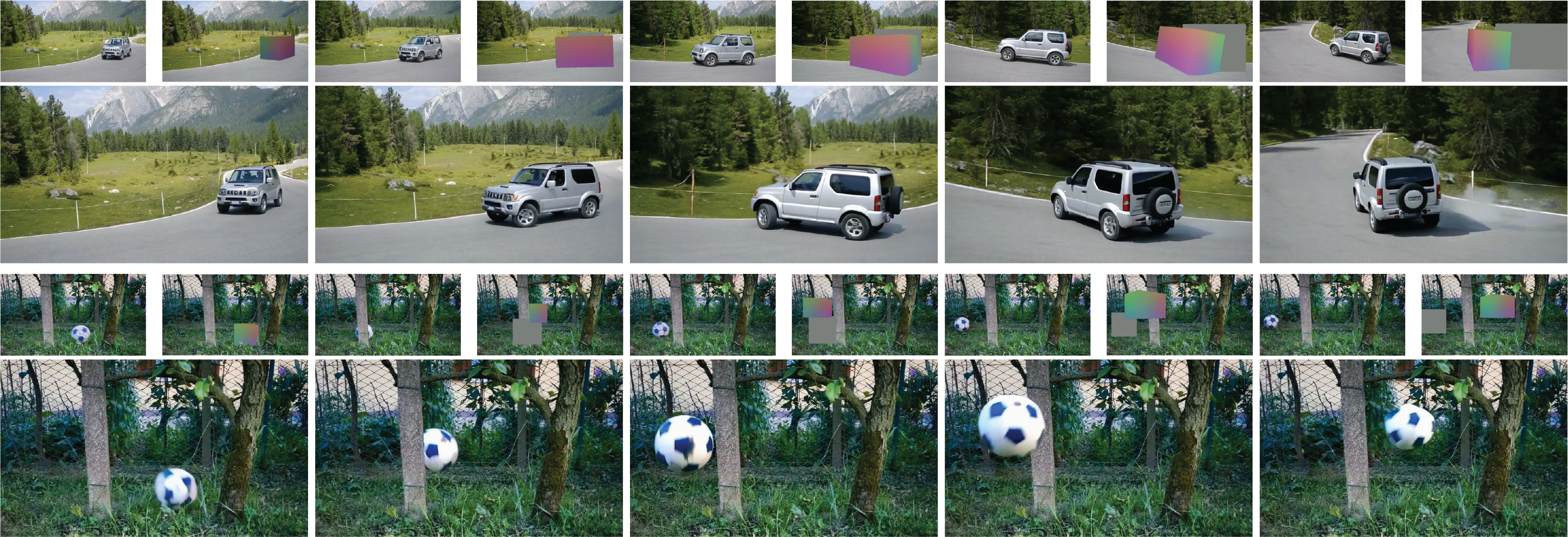} 
    \caption{\textbf{Video motion editing via oriented 3D proxy manipulation.} 
    Insets visualize the (left) original video frame and (right) our editing interface, where the source object is neutralized (gray) and the new target trajectory is specified using our hybrid DNOCs-based oriented boxes. 
    \textit{Top:} Dynamic trajectory retargeting. We modify a jeep's standard path from the DAVIS dataset~\cite{davis-Perazzi2016} into a high-speed drift. Our model successfully infers secondary physics-based effects, including realistic tire skids and volumetric smoke generated by sudden braking. 
    \textit{Bottom:} Complex occlusion and spin editing. A football's simple linear path is edited into a weaving trajectory. This requires delicate spatio-temporal reasoning to handle intricate occlusions behind trees and foliage, while simultaneously maintaining consistent spin and orientation throughout the new motion. 
    Our method, even with simple-to-author controls, enables these high-fidelity edits through sparse 3D proxies, preserving the original scene's identity while significantly altering its dynamics. Please see our supplemental webpage for video results.}
    \label{fig:edit_gallery}
\end{figure}
\section{Evaluation and Results}
\label{sec:results}
While previous work has looked at the problem of 3D camera and 2D/3D bounding box controls for video generation models, we believe that we are the first to present a holistic combination of easy-to-author 3D ``blocking'' proxies combined with the ability to author complex, multi-object interactions and dynamics. We propose several novel metrics to evaluate the quality of such proxy-controlled video generation models and we evaluate our method against several baselines.

In particular, we measure whether generated videos respect object trajectories, spatial containment, motion dynamics, and occlusion ordering defined by the control signals. We compare against baselines that rely on weaker structural representations such as 2D optical flow or 2D bounding boxes, both common form of control signals adopted in controllable video generation methods.

\paragraph{\textbf{Datasets.}}
To measure the quality of our method in real-world scenarios, we make use of real-world video datasets that already come with 3D box annotations. We evaluate on three datasets covering both real-world driving scenes and complex human-object interactions.
%
\textit{(i) nuScenes~\cite{nuscenes20}}
 is a large-scale autonomous driving dataset containing urban driving videos with annotated 3D bounding boxes for multiple dynamic objects such as vehicles, pedestrians, and cyclists. We use the validation split and derive control signals directly from ground-truth 3D bounding boxes. These boxes capture full 3D position, orientation, and scale over time, enabling evaluation of spatial grounding and trajectory fidelity in real-world scenes.
\textit{(ii) HO-3D~\cite{hampali2020ho3d}}
 contains sequences of human hand-object interactions with accurate object pose annotations and mesh models. We derive oriented bounding boxes from the ground-truth object poses and meshes. The dataset is challenging due to rapid rotations, articulated hand motion, and frequent occlusions between hands and manipulated objects.
\textit{(iii) BEHAVE~\cite{bhatnagar22behave}}
 focuses on full-body human interactions with large objects such as chairs and suitcases. Compared to HO-3D, the dataset contains more significant object motion and complex physical interactions. As in HO-3D, we derive oriented bounding boxes from ground-truth poses and meshes. Across all the datasets, the control signals are rendered into conditioning videos that are provided to the generation model.

\begin{figure}[t!]
    \centering
    \includegraphics[width=\linewidth]{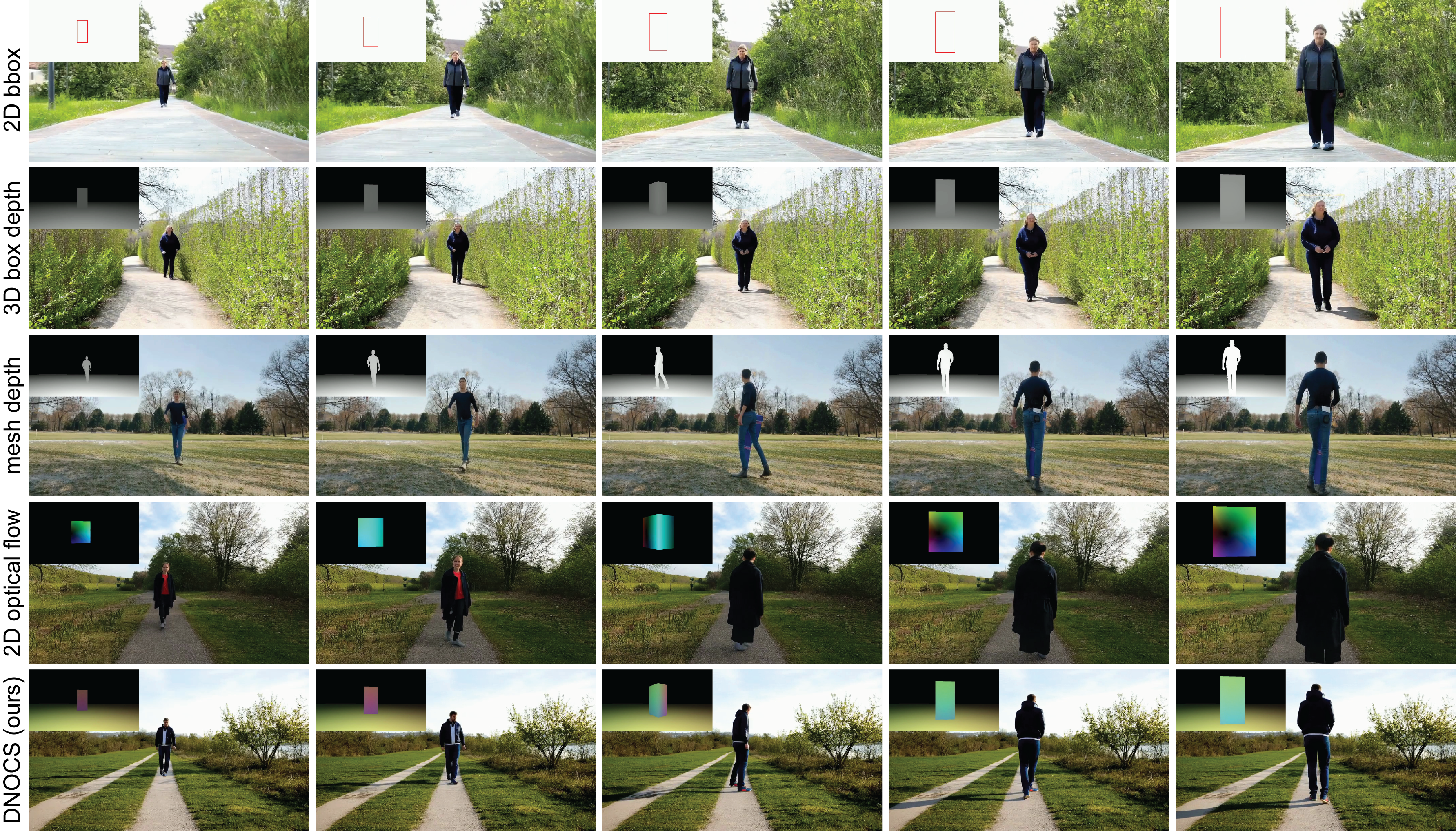}
   \caption{\textbf{Qualitative comparison.} We compare our DNOCs-based oriented box control against several alternatives, including 2D bounding boxes, 3D box depth, mesh depth, and 2D optical flow. While 2D-centric methods struggle with viewpoint-consistent orientation and temporal grounding, and dense depth/mesh guidance can over-constrain natural dynamics, our method (bottom) excels at preserving precise 6-DOF choreography (note the 360 rotation of the person) while still generating natural object deformations and realistic environmental interactions.}
    \label{fig:comparison}
\end{figure}

{\textbf{Evaluation Metrics.}}
We evaluate structural adherence using metrics based on scene flow and mask tracking via video segmentation. These evaluate spatial grounding, trajectory accuracy, motion consistency, and occlusion reasoning. 
For each object $i$ at time $t$, we denote the 2D projected control box mask as $B_{i,t}$ and the generated object mask as $M_{i,t}$, obtained via box-prompted video segmentation and tracking using SAM-2\cite{ravi2024sam}.

\paragraph{(i) Containment (Contain $\uparrow$)}
 measures how well generated object pixels remain within the prescribed control region:
\begin{equation}
\text{Contain}_{i,t} =
\frac{|M_{i,t} \cap B_{i,t}|}{|M_{i,t}| + \varepsilon}.
\end{equation}

The reported score averages over all objects and frames.

\paragraph{(ii) Trajectory Error (TrajErr $\downarrow$)}
 measures the deviation between the generated object trajectory and the control trajectory. Let $c_{i,t}$ be the centroid of the generated object mask and $b_{i,t}$ the projected center of the control box. The error is
\begin{equation}
\text{TrajErr} =
\frac{1}{NT}\sum_{i=1}^{N}\sum_{t=1}^{T}
\|c_{i,t}-b_{i,t}\|_2.
\end{equation}

\paragraph{(iii) Occlusion Accuracy (OcclAcc $\uparrow$)}
 measures whether the generated video respects correct depth ordering between overlapping objects. Let $O_{n,f,t}$ denote the overlap region between near object $n$ and far object $f$. The near dominance ratio is defined as
\begin{equation}
\text{NDR}_t =
\frac{|M_{n,t}\cap O_{n,f,t}|}{|O_{n,f,t}|+\varepsilon}.
\end{equation}
A frame is counted as correct if $\text{NDR}_t > 0.5 + \delta$, where $\delta=0.05$. Occlusion accuracy is the fraction of correct frames among all valid overlaps.

\paragraph{(iv) Rigid Motion Consistency (RMC $\downarrow$)}
 evaluates whether the generated scene motion matches the rigid motion implied by the control animation. Scene flow is estimated from predicted optical flow using Raft~\cite{teed2020raft} and depth using VDA~\cite{video_depth_anything25}, producing a 3D motion vector $v_t(x,y)$ for each pixel. Let $\Delta T_{i,t}$ denote the rigid motion of object $i$ between frames. For pixels inside the object region, RMC is defined as the error between specified rigid motion and estimated scene flow:
\begin{equation}
\text{RMC} =
\text{median}_{(x,y)\in B_{i,t}}
\|X_{t+1}(x,y)-\Delta T_{i,t}X_t(x,y)\|_2.
\end{equation}

\paragraph{(v) Global Motion Field Agreement (GMFA $\downarrow$)} 
measures agreement between the estimated scene flow $v_t(x,y)$ and the motion field predicted from the control animation (combining camera motion and object rigid motion). It is computed as the median motion error over all pixels:
\begin{equation}
\text{GMFA} =
\text{median}_{(x,y)}
\|v_t(x,y)-v_{\text{pred}}(x,y)\|_2.
\end{equation}

\paragraph{(vi) Global Overlap Winner (GOW $\uparrow$)} 
 evaluates occlusion reasoning in overlapping regions. For each pixel in the overlap region $O$, we measure how well the observed scene flow is explained by the rigid motion of the near and far objects, respectively, and calculate accuracy:
\begin{equation}
\text{GOW} =
\frac{1}{|O|}
\sum_{(x,y)\in O}
\mathds{1}[e_n(x,y) < e_f(x,y)],
\end{equation}
where $e_n(x,y) = \| v(x,y) - v_n(x,y) \|_2$ and $e_f(x,y) = \| v(x,y) - v_f(x,y) \|_2$. denote motion reconstruction errors for near and far objects, respectively.

\paragraph{(vii) Visual Quality}  using  ``overall normalized average quality'' from VBench~\cite{huang2023vbench}.

\subsection{Results}
\begin{table}[t]
\centering
\tabsmall
\setlength{\tabcolsep}{3.5pt}
\renewcommand{\arraystretch}{1.12}
\caption{\textbf{Real-world structural control on nuScenes~\cite{nuscenes20}.} We evaluate the ability of various control signals to reconstruct real-world dynamics using ground-truth 3D boxes from the \textbf{nuScenes} validation set. Ours, utilizing oriented 3D boxes, significantly outperforms 2D-based baselines on relevant metrics~\cite{huang2023vbench} in spatial grounding (TrajErr) and occlusion reasoning (OcclAcc), while preserving high visual quality. Upward arrows ($\uparrow$) indicate higher is better; \textbf{best} results are in bold, \second{second best} are underlined.
}
\label{tab:nuscenes_real3d}

\resizebox{\linewidth}{!}{%
\begin{tabular}{lcccccccc}
\toprule
Method & Input & Contain\up & GOW\up & GMFA\down & RMC\down & TrajErr\down & OcclAcc\up & Quality \\
\midrule
Control-free Baseline &  GT First + Last frames                          & 10.22 & 40.65 & 0.828 & 0.863 & 90.12 & 41.45 & \best{76.45} \\
VACE        &   2D Flow                & 21.23 & \second{86.76} & 0.135 & 0.566 & 7.86 & 73.91 & 73.90 \\
VACE ft        &   2D Flow                & 22.45 & 85.32 & \second{0.093} & \second{0.528} & \second{6.78} & \second{79.32} & \second{75.50} \\
VACE ft        &   2D Boxes          & \best{96.33} & 42.33 & 0.232 & 0.735 & 16.66 & 42.45 & 66.34 \\

\best{LooseControlVideo} &  Rendered Oriented 3D Boxes         & \second{87.93} & \best{97.32} & \best{0.066} & \best{0.318} & \best{5.79} & \best{92.69} & 74.45 \\
\bottomrule
\end{tabular}%
}
\end{table}
\begin{table}[t]
\centering
\tabsmall
\setlength{\tabcolsep}{3.5pt}
\renewcommand{\arraystretch}{1.12}
\caption{
\textbf{Comparison of object-centric structural control on HO-3D~\cite{hampali2020ho3d} and BEHAVE~\cite{bhatnagar22behave}.} We evaluate our method on complex human/object interactions by deriving 3D oriented bounding boxes from ground-truth poses and meshes. Our approach (\name~\textbf{LCV}) consistently achieves superior motion alignment (GMFA, RMC) and trajectory accuracy (TrajErr), demonstrating that oriented 3D primitives provide a more robust prior for complex/deforming motion (e.g., involving spings and turns) and close-contact interactions than 2D flow or axis-aligned boxes. ``ft" implies finetuning on our training set.
}
\label{tab:objectcentric}

\resizebox{\linewidth}{!}{%
\begin{tabular}{llcccccccc}
\toprule
Dataset & Method & Input & Contain$\uparrow$ & GOW$\uparrow$ & GMFA$\downarrow$ & RMC$\downarrow$ & TrajErr$\downarrow$ & OcclAcc$\uparrow$ & Quality \\
\midrule

\multirow{5}{*}{HO-3D}
& Control-free & GT First + Last 
& 46.8 & 52.1 & 0.440 & 0.362 & 38.5 & 53.4 & \best{76.4} \\

& VACE & 2D Flow 
& 63.4 & \second{89.7} & 0.083 & 0.214 & 6.3 & 81.5 & 73.3 \\

& VACE ft & 2D Flow 
& 69.1 & 88.2 & \second{0.071} & 0.192 & \second{5.4} & \second{84.2} & \second{73.6} \\

& VACE ft & 2D Boxes 
& \best{97.9} & 56.8 & 0.126 & \second{0.181} & 9.7 & 55.1 & 72.4 \\

& \best{LCV (Ours)} & Rendered 3D Boxes 
& \second{91.3} & \best{97.4} & \best{0.045} & \best{0.122} & \best{3.9} & \best{94.1} & 72.9 \\

\midrule

\multirow{5}{*}{BEHAVE}
& Control-free & GT First + Last 
& 42.3 & 46.9 & 0.611 & 0.490 & 54.8 & 48.6 & \best{76.2} \\

& VACE & 2D Flow 
& 57.2 & \second{86.3} & 0.121 & \second{0.341} & 8.9 & 74.8 & 75.1 \\

& VACE ft & 2D Flow 
& 63.8 & 84.7 & \second{0.098} & 0.318 & \second{7.6} & \second{78.5} & \second{75.8} \\

& VACE ft & 2D Boxes 
& \best{95.8} & 48.2 & 0.238 & 0.412 & 14.9 & 49.7 & 69.3 \\

& \best{LCV (Ours)} & Rendered 3D Boxes 
& \second{88.6} & \best{95.6} & \best{0.062} & \best{0.207} & \best{5.8} & \best{90.2} & 75.0 \\

\bottomrule
\end{tabular}%
}
\end{table}
\vspace*{-.1in}
Table~\ref{tab:nuscenes_real3d} presents results on the \textbf{nuScenes} dataset. Our method significantly improves structural adherence across key metrics. In particular, it achieves the lowest trajectory error and highest occlusion accuracy, indicating that oriented 3D boxes rendered in DNOCS provide strong geometric priors for modeling real-world scene dynamics. Compared to 2D flow-based baselines, which encode motion only in image space, our representation captures full 3D spatial relationships including orientation and depth ordering.
The 2D box baseline achieves high containment scores due to tight spatial constraints in image space, but lacks orientation and depth information, resulting in worse motion alignment and trajectory accuracy. In contrast, our method maintains strong containment while significantly improving motion fidelity.

Table~\ref{tab:objectcentric} reports results on \textbf{HO-3D} and \textbf{BEHAVE}, datasets involving complex human-object interactions with rotations and close-contact motion. Our method consistently achieves the best performance in motion alignment (GMFA, RMC) and trajectory accuracy (TrajErr). These improvements demonstrate that oriented 3D primitives provide a stronger structural prior for complex object motion than 2D flow or boxes. The control-free baseline achieves the highest perceptual quality by receiving ground-truth frames but fails to reproduce correct motion, resulting in large trajectory errors. Overall, our approach provides the best balance between structural fidelity and visual realism.

We present qualitative results for video generation in \Cref{fig:results_gallery}. As can be seen here, \name can be used to author extremely complex video stories with intricate object dynamics. This includes trajectories that converge to create interactions (e.g., the eagle and rabbit in \Cref{fig:teaser} top), trajectories are designed to \textit{avoid} interactions (the weaving car in \Cref{fig:results_gallery} top), objects that disappear and re-appear because of occlusions (the cat in \Cref{fig:results_gallery} middle) and scenes where orientations can convey deeply meaningful narratives (the puppies facing each other and playing in \Cref{fig:results_gallery} bottom). In all these cases, \name allows for a simple, expressive control authoring interface that leverages video models to generate complex appearance, animations, and secondary dynamics.

\Cref{fig:edit_gallery} presents results for video \textit{editing} and demonstrates that \name can edit the motion of individual objects in the scene while preserving the overall identity and appearance of this scene, generating secondary physics-based effects (the skidding car in \Cref{fig:edit_gallery} top) and staying faithful to scene geometry and occlusions (the soccer ball in \Cref{fig:edit_gallery}). This makes \name a powerful tool for post-production video editing. 

In \Cref{fig:comparison}, we evaluate alternate control signals---specifically 2D bounding boxes, dense 2D depth maps constructed from 3D bounding boxes or 3D meshes, and 2D optical flow. Our spatial blocking paradigm combined with the DNOCS-based control signal does the best job of capturing object motions with view and motion ambiguities (an issue with 2D centric methods) while preserving the flexibility to infer natural motion and secondary effects (a challenge with dense guidance methods that, additionally, are challenging to author).

\subsection{Limitations}

\paragraph{Identity-to-Box Assignment.} Our primary limitation is that we do not \textit{explicitly} bind specific visual identities to individual 3D boxes. This binding problem is particularly challenging in a residual ControlNet-style architecture where global features often blend. In the future, we aim to incorporate multi-view reference images assigned to specific 3D anchors \cite{saha2025sigmagen}, allowing for consistent identity preservation in complex, multi-character scenes.

\textit{Manual Timing.} Although our method decouples deformation from trajectory, the user is still responsible for authoring the temporal rhythm of interactions. While this aligns with professional animator workflows which prioritize precise control over motion timing, an interesting future direction is to be able to specify spatial trajectories sparsely and let the generative model infer timing.

\section{Conclusion}

We have introduced \name as a new simple-but-effective control mechanism that bridges the gap between high-level creative intent and low-level video synthesis through the use of \textit{oriented 3D bounding boxes}. By adopting the \textit{blocking metaphor} commonly used in  cinematography and game design, we demonstrate that sparse, 3D-aware primitives provide a sufficient yet loose geometric prior to orchestrate complex multi-object events -- effectively decoupling global choreography (aka timing) from fine-grained object dynamics (aka deformations). Our experiments across diverse datasets, including driving scenes~\cite{nuscenes20} and articulated human-object interactions~\cite{bhatnagar22behave}, show that \name achieves better spatial adherence and temporal consistency compared to traditional 2D-based control signals using video metrics~\cite{huang2023vbench}. Ultimately, this approach moves us closer to an intuitive, simple-to-use, director-in-the-loop paradigm for video generation, where complex physical interactions can be authored with ease without sacrificing photorealistic fidelity or structural rigor.

\bibliographystyle{splncs04}
\bibliography{main,content/ref_disocclusion}
\clearpage
\begin{center}
    {\LARGE Supplementary Material}\\[1em]
\end{center}

\vspace{1cm}

\appendix
\renewcommand{\thetable}{A\arabic{table}}
\setcounter{table}{0}

\section{User Study}

\begin{table}[ht]
\centering
\small
\caption{Overall completed-session pairwise preference matrix. Each cell reports
the percentage of votes preferring the row method over the column method
(64 votes per method pair).}
\label{tab:user_study_overall}
\begin{tabular}{lcccc}
\hline
Method & LCV (Ours) & Depth Only & Optical Flow & 2D Boxes \\
\hline
LCV (Ours) & -- & 78.1\% & 87.5\% & 92.2\% \\
Depth Only & 21.9\% & -- & 65.6\% & 75.0\% \\
Optical Flow & 12.5\% & 34.4\% & -- & 50.0\% \\
2D Bounding Boxes & 7.8\% & 25.0\% & 50.0\% & -- \\
\hline
\end{tabular}
\end{table}

We conduct a user study to evaluate the perceptual preference of generated videos under different control signals. 
In particular, we compare our method (\textbf{LCV}) against three baselines that use weaker structural controls: \textbf{Depth Only}, \textbf{Optical Flow}, and \textbf{2D Bounding Boxes}. 
The goal of the study is to measure which method produces videos that better follow the intended object motion while remaining visually plausible.

\paragraph{Study Setup.}
We perform a pairwise comparison study using a two-alternative forced choice (2AFC) protocol. 
For each test case, participants are shown the conditioning input video (rendered control signal) and the original video (in case of editing), together with two generated videos produced by different methods. 
Participants are asked to select the video that better follows the intended input motion while maintaining overall visual realism. 
Each generated video is approximately 5 seconds long.

The study contains multiple test scenes covering both \emph{generation} and \emph{editing} scenarios. 
In generation cases, the conditioning video contains only rendered control signals. 
In editing cases, participants are additionally shown the original input video to provide context for the intended edit.

A total of 16 participants completed the study. 
For each method pair we collect 64 votes across all test cases.

\paragraph{Overall Results.}
Table~\ref{tab:user_study_overall} reports the overall pairwise preference matrix across all cases. 
Each cell shows the percentage of votes preferring the method in the row over the method in the column. 
Our method (\textbf{LCV}) is strongly preferred over all baselines, winning 78.1\%, 87.5\%, and 92.2\% of comparisons against Depth Only, Optical Flow, and 2D Boxes respectively. 
These results indicate that participants consistently favor videos generated using oriented 3D box controls.

\paragraph{Editing Results.}
Table~\ref{tab:user_study_editing} reports results for editing scenarios. 
Our method again receives the highest preference rates, winning 84.4\%, 90.6\%, and 84.4\% of comparisons against Depth Only, Optical Flow, and 2D Boxes respectively. 
This suggests that oriented 3D control signals provide clearer structural guidance when modifying existing videos.

\paragraph{Generation Results.}
Table~\ref{tab:user_study_generation} shows results for pure generation scenarios. 
LCV achieves strong preference across all comparisons, winning 71.9\% against Depth Only, 84.4\% against Optical Flow, and 100\% against 2D Boxes. 
These results demonstrate that the richer 3D structural representation of DNOCS leads to outputs that users perceive as better aligned with the intended motion.

Overall, the user study confirms that participants consistently prefer videos generated using our DNOCS representation over methods relying on weaker control signals.

\begin{table}[ht]
\centering
\small
\caption{Editing-only completed-session pairwise preference matrix. Each cell
reports the percentage of votes preferring the row method over the column
method (64 votes per method pair).}
\label{tab:user_study_editing}
\begin{tabular}{lcccc}
\hline
Method & LCV (Ours) & Depth Only & Optical Flow & 2D Boxes \\
\hline
LCV (Ours) & -- & 84.4\% & 90.6\% & 84.4\% \\
Depth Only & 15.6\% & -- & 90.6\% & 59.4\% \\
Optical Flow & 9.4\% & 9.4\% & -- & 12.5\% \\
2D Bounding Boxes & 15.6\% & 40.6\% & 87.5\% & -- \\
\hline
\end{tabular}
\end{table}

\begin{table}
\centering
\small
\caption{Generation-only completed-session pairwise preference matrix. Each
cell reports the percentage of votes preferring the row method over the column
method (64 votes per method pair).}
\label{tab:user_study_generation}
\begin{tabular}{lcccc}
\hline
Method & LCV (Ours) & Depth Only & Optical Flow & 2D Boxes \\
\hline
LCV (Ours) & -- & 71.9\% & 84.4\% & 100.0\% \\
Depth Only & 28.1\% & -- & 40.6\% & 90.6\% \\
Optical Flow & 15.6\% & 59.4\% & -- & 87.5\% \\
2D Bounding Boxes & 0.0\% & 9.4\% & 12.5\% & -- \\
\hline
\end{tabular}
\end{table}

\end{document}